\documentclass[runningheads]{llncs}

\usepackage[T1]{fontenc}
\usepackage{graphicx}
\usepackage{booktabs}       
\usepackage{multirow}
\usepackage{array}
\usepackage{xcolor}
\usepackage{hyperref}
\usepackage{amsmath}
\usepackage{amssymb}
\usepackage{microtype}      

\definecolor{myred}{RGB}{192,0,0}
\definecolor{mygreen}{RGB}{0,128,0}
\definecolor{mygray}{RGB}{128,128,128}

\newcommand{\yes}{\textcolor{mygreen}{\checkmark}}
\newcommand{\no}{\textcolor{myred}{$\times$}}
\newcommand{\partialmark}{\textcolor{orange}{$\sim$}}

\begin{document}

\title{Metrics or Mirage? An Audit of Evaluation Inconsistencies
       in Colonoscopy Polyp Segmentation Benchmarks}

\author{Aisha Urooj\inst{1} \and
Zain Ul Abdien\inst{2} \and
Neelu Madan\inst{3}}

\institute{
Lunit, South Korea \quad \and
RPTU Kaiserslautern, Germany \quad \and
Aalborg University \& Pioneer Center, Denmark
}

\maketitle

\begin{abstract}
Progress in colonoscopy polyp segmentation is routinely reported
through leaderboard comparisons on a small set of public benchmarks.
We argue that this apparent progress is difficult to verify:
a systematic audit of \textbf{27 papers} published between 2015 and 2026
reveals three structural problems in how the community evaluates models.
\textbf{First}, 25 of 27 papers \textit{omit the Hausdorff distance}. Hausdorff distance is a boundary-accuracy metric with direct clinical relevance for detecting flat or small polyps, and is a standard in radiotherapy segmentation.
\textbf{Second}, at least five \textit{incompatible train/test split protocols} co-exist across papers reporting results on the same two datasets (Kvasir-SEG and CVC-ClinicDB), making published Dice scores non-comparable even when they appear in the same leaderboard column.
\textbf{Third}, 26 of 27 papers make \textit{performance claims without any statistical significance test}.
Strikingly, four papers published \emph{after} the Metrics Reloaded
framework~\cite{metricsreloaded2024} (Maier-Hein et al., \textit{Nature Methods} 2024)
perpetuate these same problems, suggesting that general-purpose
metric guidance has not yet reached the colonoscopy sub-community.
To show these problems are not merely cosmetic, we re-evaluate five
representative models under three controlled protocols with a single
uniform scorer, and find that the reported metric conceals large boundary
and recall failures, that the ``best'' model changes with the metric, and
that near-tied rankings reverse across random splits.
We propose a five-point \textbf{Polyp Segmentation Reporting Checklist}~(PSRC) as a lightweight, domain-adapted corrective.
\end{abstract}

\section{Introduction}
\label{sec:intro}

\begin{figure}[t]
\centering
\includegraphics[width=\linewidth]{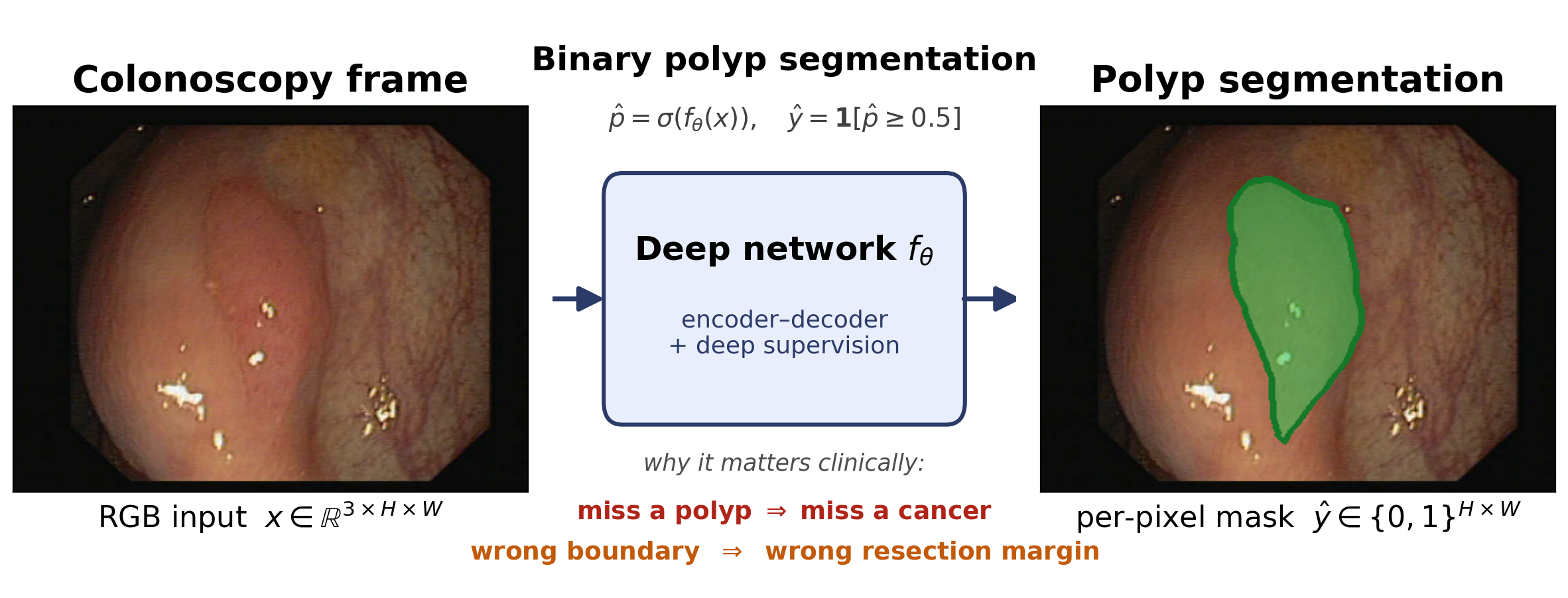}
\vspace{-0.6cm}
\caption{\textbf{The task.} Binary polyp segmentation maps a colonoscopy frame to
a per-pixel polyp mask via an encoder--decoder network with deep supervision. The
two clinically decisive failure modes are exactly the ones region-overlap metrics
under-report: a missed polyp is a missed cancer (recall), and an inaccurate
boundary is a wrong resection margin (HD95/NSD).}
\vspace{-0.6cm}
\label{fig:task}
\end{figure}

\begin{figure}[h]
\centering
\includegraphics[width=0.92\linewidth]{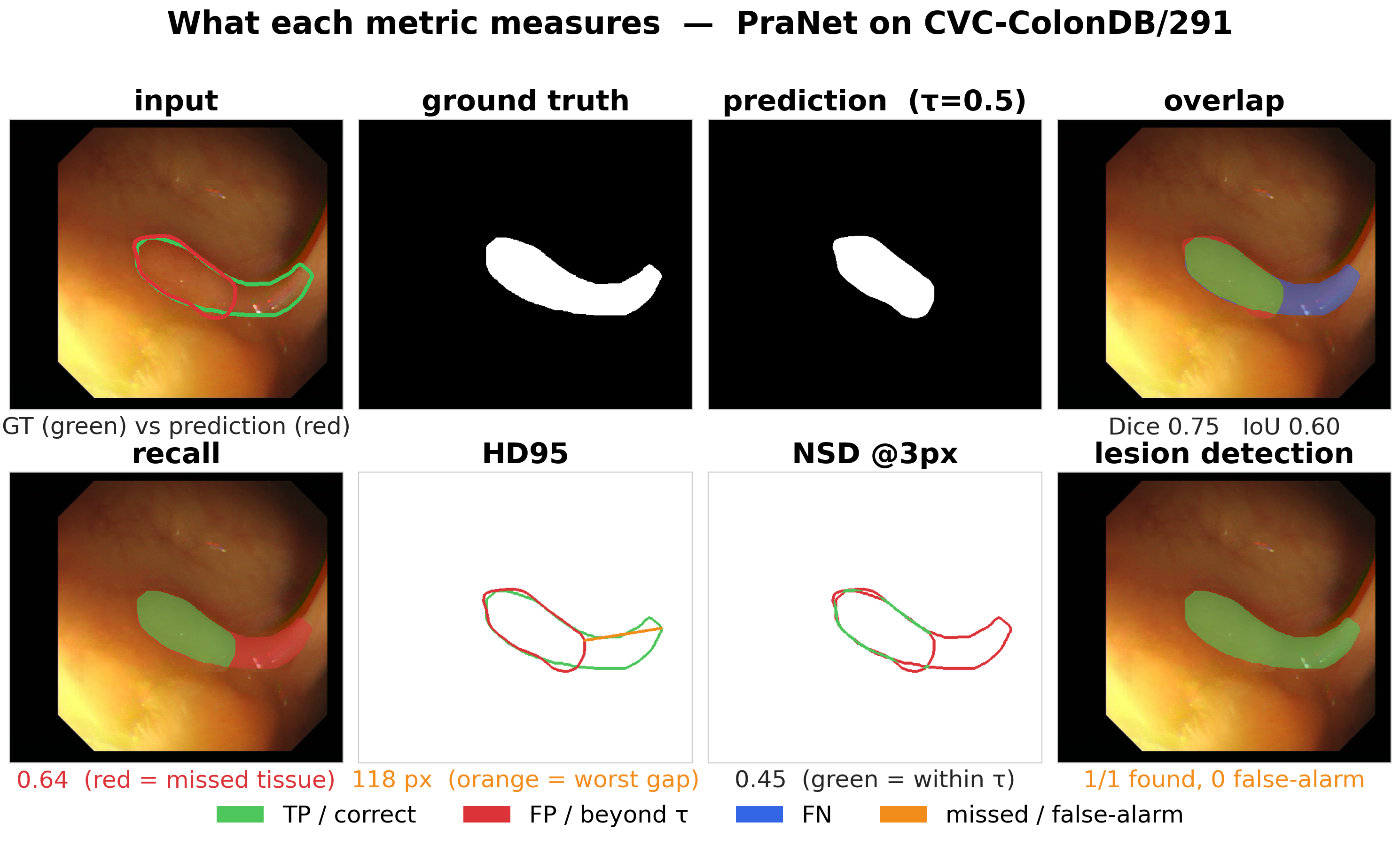}
\vspace{-0.4cm}
\caption{What each metric measures on one prediction (Dice $0.75$): the recall
panel shows $\sim$36\% of the polyp missed and HD95 a $118$\,px boundary error --
both invisible in Dice alone.}
\label{fig:explainer}
\vspace{-0.4cm}
\end{figure}
Colorectal cancer is the third most prevalent cancer worldwide, and colonoscopy remains the primary screening modality~\cite{pranet2020}.
However, manual examination is highly prone to human error, with up to 20-25\% of precancerous polyps missed during routine procedures due to visual fatigue or subtle lesion morphology~\cite{rex1997colonoscopic,ZHAO20191661}.
Because early detection and removal of these adenomas significantly lowers mortality rates, automated polyp segmentation can assist endoscopists by highlighting lesions in
real time.
Consequently, the past five years have seen a rapid succession of deep learning models that each claim state-of-the-art (SOTA) performance on standard benchmarks.

Yet a closer inspection of how SOTA is evaluated reveals a troubling pattern. The field converged around a single evaluation template introduced by PraNet~\cite{pranet2020} in 2020, where six metrics are reported across five datasets using a fixed train/test split. The metrics used in this template are: Mean Dice, Mean IoU, Weighted F-Measures, Structure Measure, Mean Absolute Error, and Enhanced-alignment Measure. The datasets used in the template are: Kvasir-SEG \cite{jha2019kvasir}, CVC-ClinicDB/CVC-612~\cite{bernal2015wm}, CVC-ColonDB~\cite{tajbakhsh2015automated}, ETIS-LaribPolypDB~\cite{silva2014toward}, EndoScene~\cite{vazquez2017benchmark}
This template is then subsequently copied, sometimes verbatim, by more than a dozen
subsequent papers (e.g., ColonFormer~\cite{colonformer2022}, Polyp-PVT~\cite{polypvt2023}, SA-Net~\cite{sanet2021}) without questioning whether those six metrics adequately capture what matters clinically.

From a clinical standpoint, pixel-level overlap metrics like Mean Dice and Mean IoU remain inherently detached from the core diagnostic objectives of gastroenterologists. In live endoscopy, a model's true utility is governed by its lesion-level sensitivity (avoiding catastrophic false negatives) and its robustness to Out-of-Distribution (OOD) real-world shifts. Standard evaluation templates completely fail to capture how an architecture handles covariate shifts typical of routine clinical practice, such as variation in camera sensors, localized lighting artifacts, poor bowel preparation, or obscure mucosal debris.

Meanwhile, a parallel line of work in general medical image analysis has documented that metric choice is far from neutral. The Metrics Reloaded framework~\cite{metricsreloaded2024} demonstrated
that ``chosen performance metrics often do not reflect domain interest,''
and identified Hausdorff Distance (HD95) as a critical complement to overlap metrics whenever boundary accuracy is clinically relevant.
For polyp segmentation, where missed flat or sub-centimetre lesions
carry life-or-death consequences, this gap is not merely academic. This paper makes four contributions:
\begin{enumerate}
  \item A systematic audit of 27 polyp segmentation papers (2015--2026), characterising their evaluation practices across four dimensions, namely: \textit{Metric Profile}, \textit{Data Partitioning}, \textit{Generalization Scope}, and \textit{Statistical Rigor}.
  \item We empirically show that incompatible split protocols make direct numeric comparisons invalid across most published works.
  \item A controlled re-evaluation of five representative models under three protocols (fixed split, random split, and out-of-distribution), reporting the omitted metrics (boundary, recall, lesion-level) with per-image significance, showing that near-tied rankings are not robust and that metric choice changes the winner.
  \item The Polyp Segmentation Reporting Checklist (PSRC), a five-point domain-adapted standard for future papers.
\end{enumerate}

\section{Related Work}
\label{sec:related}

\begin{figure}[t]
\centering
\includegraphics[width=\linewidth]{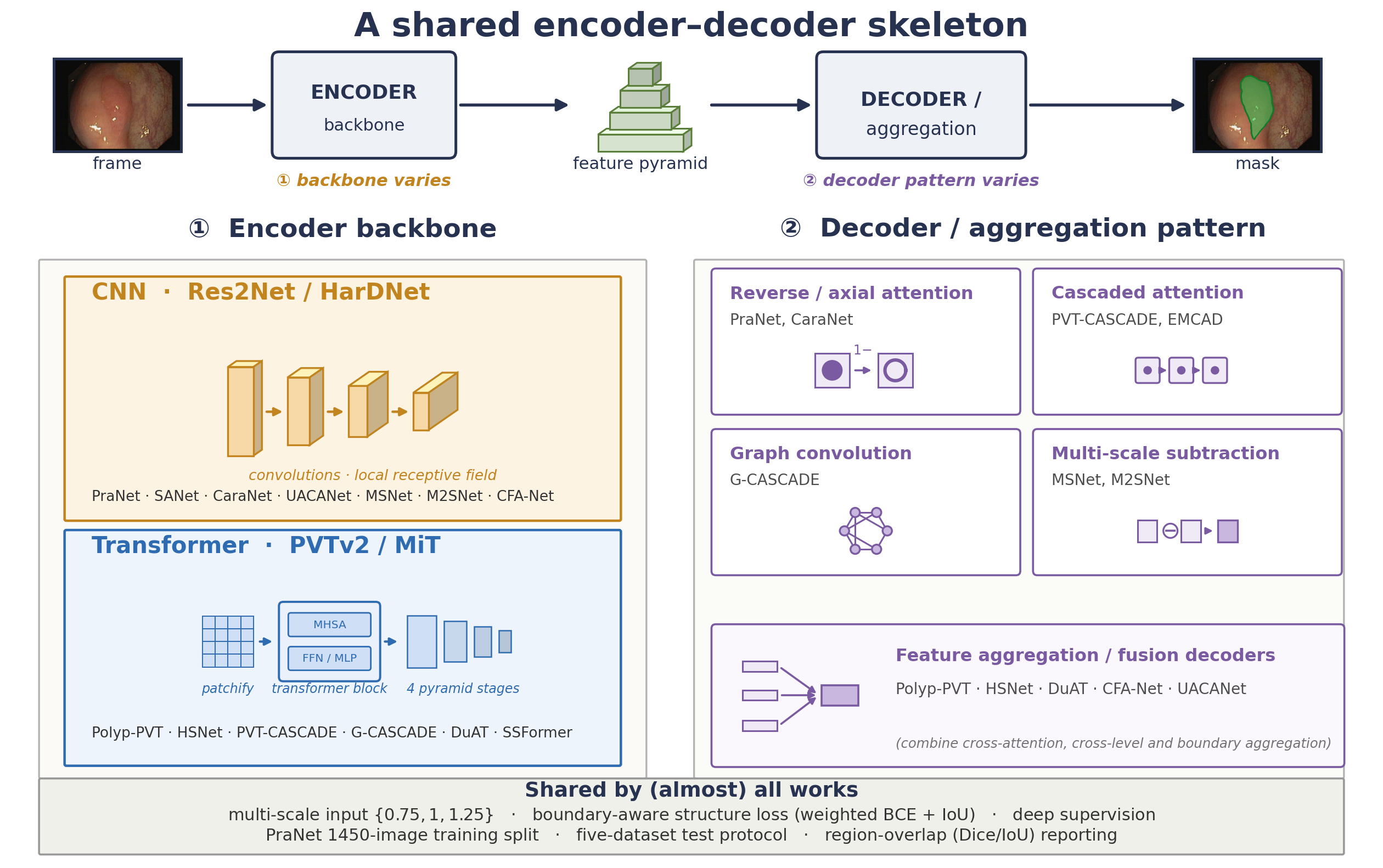}
\vspace{-0.4cm}
\caption{\textbf{Architectural landscape of the studied works.} Despite steady
reported gains, the models share one encoder--decoder skeleton and differ only
along two axes: the \emph{encoder backbone} (CNN Res2Net/HarDNet vs.\ transformer
PVTv2/MiT) and the \emph{decoder / feature-aggregation pattern}. Critically, they
also share their \emph{evaluation}: the same multi-scale training,
boundary-aware structure loss, deep supervision, fixed split, and five-dataset
region-overlap protocol.}
\label{fig:arch}
\vspace{-0.4cm}
\end{figure}

\noindent \textbf{Polyp segmentation architectures.}
Deep polyp segmentation spans two main eras: CNN-based attention networks and Transformer-driven multi-scale decoders. The CNN era was anchored by PraNet~\cite{pranet2020}, which introduced parallel reverse attention for boundary refinement. Subsequent CNNs targeted specific edge cases: background bias~\cite{sanet2021}, boundary uncertainty~\cite{uacanet2021}, small lesions~\cite{caranet2022}, and camouflaged targets~\cite{c2fnet2021}. With the shift to Pyramid Vision Transformer (PVT) encoders, innovation pivoted to multi-scale decoder engineering. This includes hybrid CNN-transformer fusion~\cite{hsnet2022,polypvt2023}, global-local aggregation~\cite{duat2023,cfanet2023}, feature subtraction units~\cite{m2snet2023}, and advanced attention or graph decoding heads~\cite{pvtcascade2023,gcascade2024}. While these architectures report steady mean-Dice gains, their evaluation remains bound to historical protocols, leaving clinical robustness unexamined.

\noindent\textbf{Metric selection and validation.}
The biomedical community has repeatedly warned that overlap metrics alone are
insufficient. Metrics Reloaded~\cite{metricsreloaded2024} introduces a
\emph{problem-fingerprint} framework that recommends, for a given task, a
principled metric set (e.g.\ pairing an overlap metric with the Normalized
Surface Distance for boundary-sensitive segmentation, and detection metrics when
object-level recovery matters); its companion catalogues common metric
pitfalls~\cite{reinke2024pitfalls}. Outside medicine, controlled re-evaluations
such as ``A Metric Learning Reality Check''~\cite{musgrave2020} showed that
reported progress can evaporate under a common protocol. Our work applies this
auditing logic to polyp segmentation: we adopt the Metrics Reloaded
recommendation, validate our implementations against its reference code, and
quantify how far the field's practice diverges from it.

\noindent\textbf{Benchmark reproducibility and split protocols.}
Several works note that evaluation choices, not just architectures, drive
reported numbers. Fitzgerald et al.~\cite{fcbswinv2} demonstrate that
random $80/20$ partitions of Kvasir-SEG alone shift Dice by $>1\%$, and that
seeds and loaders are rarely reported in enough detail to reproduce; yet
incompatible split protocols (PraNet-fixed vs.\ random) continue to be compared
on shared leaderboards. We extend this from a single-model observation to a
controlled multi-model re-evaluation, and additionally show that even metric
\emph{implementations} disagree (HD95 differs by up to $16$\,px between widely
used libraries), and that secondary comparison tables contain transcription
errors exceeding claimed improvements.

\noindent\textbf{Out-of-distribution and multi-center evaluation.}
Because in-distribution test sets gradually become pseudo-training signal through
repeated benchmarking, truly held-out, multi-center data is needed to assess
generalization. PolypGen~\cite{ali2023multi} provides six-center data for this
purpose, and TransNetR~\cite{transnetr2023} was among the first to evaluate
out-of-distribution. We incorporate PolypGen as a dedicated OOD protocol (P3) and
report per-center behavior, where our results show that all models degrade sharply in
absolute terms even though the coarse ranking is largely preserved.

\section{Technical Preliminaries}
\label{sec:prelim}

\noindent \textbf{Problem formulation.}
We study \emph{binary polyp segmentation}: given a colonoscopy frame, the task is to predict a pixel-wise mask delineating polyp tissue from background mucosa. Let $x \in \mathbb{R}^{3\times H\times W}$ be an RGB frame and $y \in \{0,1\}^{H\times W}$ its expert-annotated binary ground-truth mask, where $y_{ij}=1$ if pixel $(i,j)$ contains polyp tissue (figure~\ref{fig:task}). A segmentation network $f_\theta$ maps a resized input image (standardized to $352\times352$) to a logit map $\hat{s}=f_\theta(x)$. Continuous predicted probabilities are obtained via a sigmoid activation, $\hat{p}=\sigma(\hat{s})$, which are converted to a hard binary mask $\hat{y}=\mathbb{1}[\hat{p}\ge t]$ using a threshold of $t=0.5$. Prior to metric evaluation, $\hat{y}$ is bilinearly upsampled back to the frame's native resolution $(H \times W)$. The studied architectures share an encoder--decoder topology (Res2Net CNN backbones for older baselines; PVTv2 transformers for modern iterations) and employ deep supervision, emitting a set of intermediate decoder side-outputs $\{\hat{s}^{(k)}\}_{k=1}^{K}$.

\noindent \textbf{Training objectives.}
The majority of Res2Net- and PVT-based baselines are optimized using a boundary-weighted structure loss ($\mathcal{L}_{\mathrm{str}}$), which combines weighted binary cross-entropy ($\mathcal{L}_{\mathrm{wbce}}$) and weighted Intersection-over-Union ($\mathcal{L}_{\mathrm{wiou}}$) terms. To encourage sharp contours, a pixel-wise weight map $w$ is first computed based on the ground-truth target mask $y$ to heavily penalize errors along boundaries and small structures (up to a $6\times$ scaling factor):
\begin{equation}
w = 1 + 5\,\bigl|\,\mathrm{avgpool}_{31}(y) - y\,\bigr|
\end{equation}
Using this weight map, the individual loss components balance the raw network logit maps $\hat{s}$ and the sigmoid-activated probability maps $\hat{p} = \sigma(\hat{s})$ against the target mask $y$:
\begin{align}
\mathcal{L}_{\mathrm{wbce}} &= \frac{\sum w\cdot \mathrm{BCE}(\hat{s},y)}{\sum w}, \\
\mathcal{L}_{\mathrm{wiou}} &= 1 - \frac{\sum w\,(\hat{p}\,y)+1}{\sum w\,(\hat{p}+y-\hat{p}\,y)+1}.
\end{align}
The overall structure loss is then formulated as the sum of these two components:
\begin{equation}
\mathcal{L}_{\mathrm{str}} = \mathcal{L}_{\mathrm{wbce}} + \mathcal{L}_{\mathrm{wiou}}
\end{equation}
Under deep supervision, the final training objective $\mathcal{L}$ is aggregated across the logit maps $\hat{s}^{(k)}$ of all decoder side-output scales $k$:
\begin{equation}
\mathcal{L} = \sum_{k}\mathcal{L}_{\mathrm{str}}(\hat{s}^{(k)},y)
\end{equation}
which is typically optimized over a multi-scale training schedule $s \in \{0.75, 1.0, 1.25\}$. Concurrently, architectures like SANet~\cite{sanet2021} replace this structure loss with a standard cross-entropy plus Dice objective coupled with a test-time probability-correction step.
Crucially, while these dominant training objectives are explicitly engineered to be \emph{boundary-aware} ($w$) and \emph{region-aware} ($\mathcal{L}_{\mathrm{wiou}}$), downstream literature heavily favors reporting region overlap alone (Dice/IoU). This stark asymmetry between optimization properties and evaluation criteria directly motivates our boundary- and recall-oriented audit.

\noindent \textbf{Evaluation metrics.}
Supervision masks are densely hand-labeled by clinical experts~\cite{jha2019kvasir,bernal2015wm,tajbakhsh2015automated,silva2014toward,vazquez2017benchmark}. Following historical benchmark practices, primary region overlap performance is quantified via mean Dice and mean Intersection over Union (mIoU), where $\hat{y}$ denotes the predicted hard binary mask and $y$ represents the expert-annotated ground-truth target:
\begin{equation}
\mathrm{Dice}=\frac{2|\hat{y}\cap y|}{|\hat{y}|+|y|}, \qquad \mathrm{IoU}=\frac{|\hat{y}\cap y|}{|\hat{y}\cup y|}
\end{equation}
To align evaluation with the \emph{problem fingerprint} defined by the Metrics Reloaded framework~\cite{metricsreloaded2024}, we complement these region metrics with the Normalized Surface Distance (NSD@$\tau$) to capture boundary-level accuracy, utilizing the 95th percentile Hausdorff Distance (HD95) as an additional boundary metric. We further report pixel-level \emph{recall} (sensitivity) alongside a \emph{lesion-level} object detection metric determined via connected-component matching. Non-polyp (empty mask) frames are handled explicitly: total agreement between empty predicted and true masks yields a perfect score, whereas distance-based metrics (NSD, HD95) are undefined for empty targets and are systematically excluded from the corresponding averages.

\section{Audit Methodology and Meta-Dataset}
\label{sec:method}

\noindent \textbf{The PraNet baseline template.}
To audit the state of the field, we first formalize the evaluation framework established by PraNet~\cite{pranet2020}, which has served as the de facto benchmarking standard for the entire sub-field. This template dictates training on a combined set of 900 Kvasir-SEG~\cite{jha2019kvasir} and 550 CVC-ClinicDB~\cite{bernal2015wm} images (1,450 total). Models are then evaluated across the remaining images from those two datasets alongside three completely held-out sets: CVC-ColonDB~\cite{tajbakhsh2015automated}, ETIS-LaribPolypDB~\cite{silva2014toward}, and CVC-300~\cite{vazquez2017benchmark}. 

Performance within this template is traditionally distilled into a six-metric table reporting: mean Dice (mDice), mean IoU (mIoU), weighted F-measure ($F^w_\beta$), S-measure ($S_\alpha$), E-measure ($E_\phi$), and mean absolute error (MAE). Virtually all subsequent architectures including SANet~\cite{sanet2021}, UACANet~\cite{uacanet2021}, C2FNet~\cite{c2fnet2021}, Polyp-PVT~\cite{polypvt2023}, CaraNet~\cite{caranet2022}, HSNet~\cite{hsnet2022}, DuAT~\cite{duat2023}, ESFPNet~\cite{esfpnet2023}, CFA-Net~\cite{cfanet2023}, M2SNet~\cite{m2snet2023}, and G-CASCADE~\cite{gcascade2024}, adopted this template with minimal modification. This created an illusion of direct comparability across shared leaderboards, which our audit systematically stress-tests.

\noindent \textbf{Meta-dataset compilation and paper eligibility.}
To characterize benchmarking and reporting practices across this timeline, we compiled a representative cohort of $27$ deep fully-supervised polyp segmentation papers. Candidates were initially seeded from two key community references: the comprehensive survey by Mei et al.~\cite{yaosurvey2024} and the curated \emph{Awesome-Polyp-Segmentation} repository. This list was expanded via the PapersWithCode leaderboards for Kvasir-SEG~\cite{jha2019kvasir}/CVC-ClinicDB~\cite{bernal2015wm} alongside manual sweeps of recent CVPR, MICCAI, WACV, and MIDL proceedings to ensure recent 2024--2026 literature was fully represented. A work was deemed eligible for inclusion in our audit if it satisfied three strict inclusion criteria: \textbf{(a)} It introduces a novel, \emph{fully-supervised} deep architecture optimized specifically for still-image colonoscopic polyp segmentation; \textbf{(b)} It reports explicit quantitative performance values on at least one of the five core historical benchmark datasets managed under the PraNet template; \textbf{(c)} It was published, accepted, or publicly hosted on a major preprint server between January 2015 and June 2026.

To isolate the evaluation habits of mainstream model-design literature, we explicitly enforce the following exclusion criteria:
\textbf{(i)} Video polyp segmentation models whose validation protocols depend entirely on sequential video datasets (e.g., SUN-SEG~\cite{ji2022video}, CVC-VideoClinicDB~\cite{tudela2024complete});
\textbf{(ii)} Weakly-, semi-, or unsupervised segmentation frameworks that focus on learning algorithms rather than task-specific architectural design;
\textbf{(iii)} General foundation-model adaptations (e.g., zero-shot or fine-tuned SAM pipelines~\cite{Kirillov_2023_ICCV}) that do not introduce original architectural decoders;
\textbf{(iv)} Works validating exclusively on non-standard, isolated datasets (e.g., BKAI-IGH~\cite{bkai-igh-neopolyp}, PICCOLO~\cite{sanchezperalta2020piccolo}, PolypGen~\cite{ali2023multi}) that do not interface with the standard benchmarking template.
Because methodological anomalies are widespread across the resulting 27-paper sample (Table~\ref{tab:audit}), our subsequent findings describe structural reporting patterns across the entire sub-field.

\vspace{0.1cm}
\noindent \textbf{Auditing dimensions.}
For every audited paper, we systematically extract and record structural meta-data across four distinct evaluation dimensions:
\textbf{D1} \textbf{(Metric Profile):} The exact selection of reported evaluation metrics, capturing whether the work relies solely on region overlap or includes boundary-sensitive ($F^w_\beta$, $S_\alpha$, $E_\phi$, MAE, HD95, NSD) or clinical safety (Recall) parameters;
\textbf{D2} \textbf{(Data Partitioning):} The precise train/test split protocol used, noting any mixing between fixed historical partitions (e.g., PraNet-fixed) and stochastic random splits;
\textbf{D3} \textbf{(Generalization Scope):} The presence of dedicated out-of-distribution (OOD) or multi-centre external validation sets to evaluate domain shift robustness;
\textbf{D4} \textbf{(Statistical Rigor):} The reporting of statistical significance testing (e.g., $p$-values, confidence intervals) alongside absolute performance margins.

\begin{table}[h]
\vspace{-0.3cm}
\caption{Audit of evaluation practices in 27 polyp segmentation papers
(2015--2026). \yes~= reported, \no~= absent,
\partialmark~= partial / OOD on seen datasets only.
\textbf{Split codes:} P = PraNet fixed (1450 train),
R = random \%, K = Kvasir-only, C = custom, V = video.}
\label{tab:audit}
\centering
\scriptsize
\setlength{\tabcolsep}{2pt}
\begin{tabular}{llcccccccccc}
\toprule
\textbf{Paper} & \textbf{Venue (Yr)}
  & \textbf{Dice} & \textbf{HD95} & \textbf{Rec.}
  & \textbf{S-m} & \textbf{wFm} & \textbf{E-m} & \textbf{MAE}
  & \textbf{Split} & \textbf{Ext.} & \textbf{Sig.} \\
\midrule
U-Net          & MICCAI (2015)    & \yes & \no & \no & \no & \no & \no & \no & C & \no & \no \\
SFA            & MICCAI (2019)    & \yes & \no & \no & \no & \no & \no & \no & C & \no & \no \\
\midrule
PraNet         & MICCAI (2020)    & \yes & \no & \no & \yes & \yes & \yes & \yes & P & \partialmark & \no \\
ACSNet         & MICCAI (2020)    & \yes & \no & \no & \yes & \yes & \yes & \yes & P & \partialmark & \no \\
HarDNet-MSEG   & arXiv (2021)     & \yes & \no & \no & \yes & \yes & \yes & \yes & P & \partialmark & \no \\
SANet          & MICCAI (2021)    & \yes & \no & \no & \yes & \yes & \yes & \yes & P & \partialmark & \no \\
MSNet          & MICCAI (2021)    & \yes & \no & \no & \yes & \no & \yes & \yes & P & \partialmark & \no \\
UACANet        & ACM MM (2021)    & \yes & \no & \no & \yes & \yes & \yes & \yes & P & \partialmark & \no \\
C2FNet         & IEEE TMI (2021)       & \yes & \no & \no & \yes & \yes & \yes & \yes & P & \partialmark & \no \\
PNS-Net        & MICCAI (2021)    & \yes & \no & \no & \no & \no & \no & \no & V & \no & \no \\
\midrule
Polyp-PVT      & CAAI AIR (2023)  & \yes & \no & \no & \yes & \yes & \yes & \yes & P & \partialmark & \no \\
HSNet          & CBM (2022)       & \yes & \no & \no & \yes & \yes & \yes & \yes & P & \partialmark & \no \\
CaraNet        & SPIE MI (2022)   & \yes & \no & \no & \yes & \yes & \yes & \yes & P & \partialmark & \no \\
DCRNet         & ISBI (2022)      & \yes & \no & \no & \yes & \yes & \yes & \yes & P & \partialmark & \no \\
LAPFormer      & arXiv (2022)     & \yes & \no & \no & \no & \no & \no & \no & P & \partialmark & \no \\
DuAT           & PRCV (2023)      & \yes & \no & \no & \yes & \yes & \yes & \yes & P & \partialmark & \no \\
ESFPNet        & MIA (2023)       & \yes & \no & \no & \yes & \yes & \yes & \yes & P & \partialmark & \no \\
CFA-Net        & PR (2023)        & \yes & \no & \no & \yes & \yes & \yes & \yes & P & \partialmark & \no \\
M2SNet         & TPAMI (2024)     & \yes & \no & \no & \yes & \yes & \yes & \yes & P & \partialmark & \no \\
TransNetR      & MIDL (2023)      & \yes & \no & \yes & \no & \no & \no & \no & K & \yes & \no \\
FLDNet         & arXiv (2023)     & \yes & \no & \no & \yes & \yes & \yes & \yes & P & \partialmark & \no \\
IBoxCLA        & arXiv (2023)     & \yes & \no & \yes & \no & \no & \no & \no & C & \no & \no \\
\midrule
G-CASCADE      & WACV (2024)      & \yes & \no & \no & \yes & \yes & \yes & \yes & P & \partialmark & \no \\
CFFormer       & ESWA (2025)     & \yes & \yes & \no & \no & \no & \no & \no & R & \partialmark & \no \\
Polyp-DiFoM    & arXiv (2024)     & \yes & \no & \no & \no & \no & \no & \no & P & \partialmark & \no \\
HSSAM-Net      & Sci.Rep. (2025)  & \yes & \yes & \no & \no & \no & \no & \yes & C & \no & \no \\
PolyMamba-Net  & Front. (2026)    & \yes & \no & \no & \no & \no & \no & \no & R & \partialmark & \yes \\
\bottomrule
\end{tabular}
\vspace{-0.5cm}
\end{table}

\section{Literature Audit Findings}
\label{sec:audit_findings}

\noindent \textbf{Metric omissions and post-Metrics-Reloaded evolution.}
Table~\ref{tab:audit} summarizes our evaluation audit findings across all 27 representative papers spanning 2015 to 2026. The headline figures reveal a stark, systematic omission of critical evaluation metrics across the sub-field: \textbf{(i) boundary blindspots:} 25 of 27 papers completely omit HD95 or any surface-distance variant, strikingly including architectures whose primary claimed contribution is enhanced boundary delineation or edge-aware refinement (e.g., SANet~\cite{sanet2021}, CaraNet~\cite{caranet2022}); \textbf{(ii) clinical safety failures:} 25 of 27 papers fail to report Recall (Sensitivity), representing a fundamental breakdown in clinical validation reporting given that false negatives in a diagnostic screening application carry the highest clinical cost; and \textbf{(iii) absent statistical rigor:} 26 of 27 papers report no statistical significance testing whatsoever, the sole exception being PolyMamba-Net (2026), which provides Wilcoxon signed-rank $p$-values to substantiate its performance claims.

A closer look at the timeline reveals a troubling trend of metric fragmentation and regression over time. Rather than converging toward a richer, more rigorous metric set following international consensus guidelines, later papers (e.g., LAPFormer, Polyp-DiFoM) silently dropped historical summary metrics, reporting only basic Dice and IoU scores. This pattern becomes particularly acute among the four papers published well after the landmark release of the Metrics Reloaded framework by Maier-Hein et al.~\cite{metricsreloaded2024} in January 2024, i.e., G-CASCADE (WACV 2024), CFFormer (ESWA 2024), Polyp-DiFoM (arXiv 2024), and HSSAM-Net (Sci. Reports 2025): none adopted HD95 or Recall as primary evaluation metrics, and Polyp-DiFoM actually regressed to baseline Dice/IoU exclusively. This empirical timeline indicates that general-purpose biomedical metric guidance has failed to penetrate or influence the localized benchmarking inertia of this sub-community.

\noindent \textbf{Split protocol incompatibility.}
We identified five distinct train/test protocols in active simultaneous use: the \emph{PraNet fixed split} (1,450 images, used by 19 papers); an \emph{80/10/10 random split} (CFFormer 2024, PolyMamba-Net 2026); \emph{Kvasir-SEG only} (TransNetR 2023: 880/120); \emph{custom splits with extra training data} (IBoxCLA 2023 adds LDPolypVideo to training); and \emph{video splits} (PNS-Net 2021, an incomparable task). Notably, CFFormer (2024) and G-CASCADE (2024) both report Dice scores on Kvasir-SEG and CVC-ClinicDB, yet CFFormer uses an 80/10/10 random split while G-CASCADE uses the PraNet fixed split---both figures appear in comparison tables in the literature without any flag. The numerical difference in Dice attributable to split choice alone can exceed 3 percentage points~\cite{yaosurvey2024}, larger than the claimed improvement of several SOTA papers.

\noindent \textbf{External validation.}
Only one of 27 papers (TransNetR~\cite{transnetr2023}) performs genuine external validation on data from unseen centres, testing on PolypGen (six centres) and BKAI-IGH as truly unseen datasets. A further 21 report only on subsets of the five benchmark datasets that have been public for years and potentially design-influencing through repeated benchmarking, and the remaining five perform no cross-dataset evaluation at all.

\noindent \textbf{Primary source verification and transcription errors.}
To validate the numbers appearing in secondary survey tables, we retrieved Dice scores directly from the original paper PDFs for four representative models evaluated on the PraNet fixed split (Table~\ref{tab:discrepancy}); discrepancies were found in two of four papers. SANet~\cite{sanet2021} reports a ColonDB mDice of 0.753 in its original Table~1, which the survey~\cite{yaosurvey2024} copies as 0.752. More significantly, G-CASCADE~\cite{gcascade2024} reports a Kvasir-SEG mDice of 0.9274 and a CVC-ClinicDB mDice of 0.9468 in its supplementary Section~B---the only location where polyp results appear, as the main paper focuses on multi-organ segmentation, yet secondary tables report these as 0.929 and 0.936 respectively, an error of 0.011 on CVC-ClinicDB that exceeds the claimed margin of improvement of several published SOTA comparisons. We further note that G-CASCADE's polyp results being confined to supplementary material makes independent verification non-trivial, i.e., transparency issue independent of the numeric discrepancy itself.

The four-year claimed SOTA progression on Kvasir-SEG under the PraNet fixed split runs from PraNet (0.898) through SANet (0.904) and Polyp-PVT (0.917) to G-CASCADE (0.9274), a total gain of 2.94\%. Fitzgerald et al.~\cite{fcbswinv2} demonstrated empirically that different random partitions of Kvasir-SEG alone produce Dice differences exceeding 1\%, meaning split choice could account for roughly one third of the entire claimed four-year progression.

\begin{table}[t]
\caption{Transcription errors found in secondary survey tables
vs.\ primary source PDFs. The G-CASCADE CVC-ClinicDB error of
0.011 exceeds the claimed margin of improvement of several
published SOTA comparisons.}
\label{tab:discrepancy}
\centering
\small
\begin{tabular}{llccc}
\toprule
\textbf{Paper} & \textbf{Dataset}
  & \textbf{Secondary} & \textbf{Primary} & \textbf{Error} \\
\midrule
SANet~\cite{sanet2021}
  & ColonDB      & 0.752 & 0.753  & 0.001 \\
G-CASCADE~\cite{gcascade2024}
  & Kvasir-SEG   & 0.929 & 0.9274 & 0.002 \\
G-CASCADE~\cite{gcascade2024}
  & CVC-ClinicDB & 0.936 & 0.9468 & \textbf{0.011} \\
\bottomrule
\end{tabular}
\vspace{-0.4cm}
\end{table}
\section{Empirical Re-Evaluation}
\label{sec:setup}

\subsection{Framework}
We re-evaluate five representative polyp segmentation models under a single
uniform evaluator: \textbf{PraNet}~\cite{pranet2020}, \textbf{Polyp-PVT}~\cite{polypvt2023}, \textbf{SANet}~\cite{sanet2021} (released
weights), \textbf{PVT-CASCADE}~\cite{pvtcascade2023} and \textbf{G-CASCADE}~\cite{gcascade2024} (self-trained on the 1450
PraNet split, as no polyp checkpoints are released). HSNet additionally
participates in the split-sensitivity study (P2). All predictions are scored by
one implementation that computes, beyond the reported mean Dice/IoU, the omitted
\emph{boundary} metric NSD@3px (validated exactly against the official
\texttt{MetricsReloaded} package) with HD95 as a complement, and \emph{recall},
plus per-image scores for significance testing and a \emph{lesion-level}
detection metric. We evaluate under three protocols: \textbf{P1} (PraNet fixed
split, tested on the unseen CVC-ColonDB~\cite{tajbakhsh2015automated}, ETIS~\cite{silva2014toward}, CVC-300~\cite{vazquez2017benchmark}); \textbf{P2} (random
$80/20$ splits of the same pool, 3 seeds); and \textbf{P3} (out-of-distribution
on the six unseen centres of PolypGen). The released-weight models reproduce
published Dice within $\approx1\%$; the self-trained PVT-CASCADE~\cite{pvtcascade2023} and G-CASCADE~\cite{gcascade2024}
reproduce their published in-distribution Dice within $0.011$ and
$0.018$--$0.035$ respectively, confirming faithful training so their lower
unseen-set scores are genuine generalization. Metrics were verified against
\texttt{medpy} and the official \texttt{MetricsReloaded} package
(Dice/IoU/Recall/NSD exact), and a perceptual-hash$\rightarrow$SSIM audit
confirmed train/test disjointness (no leakage; max cross-set SSIM $0.66$). P2 models are trained to a fixed budget
and the \emph{final} checkpoint is reported, removing the test-set-based
checkpoint selection present in several public training scripts.

\begin{figure}[h]
\centering
\includegraphics[width=0.85\linewidth]{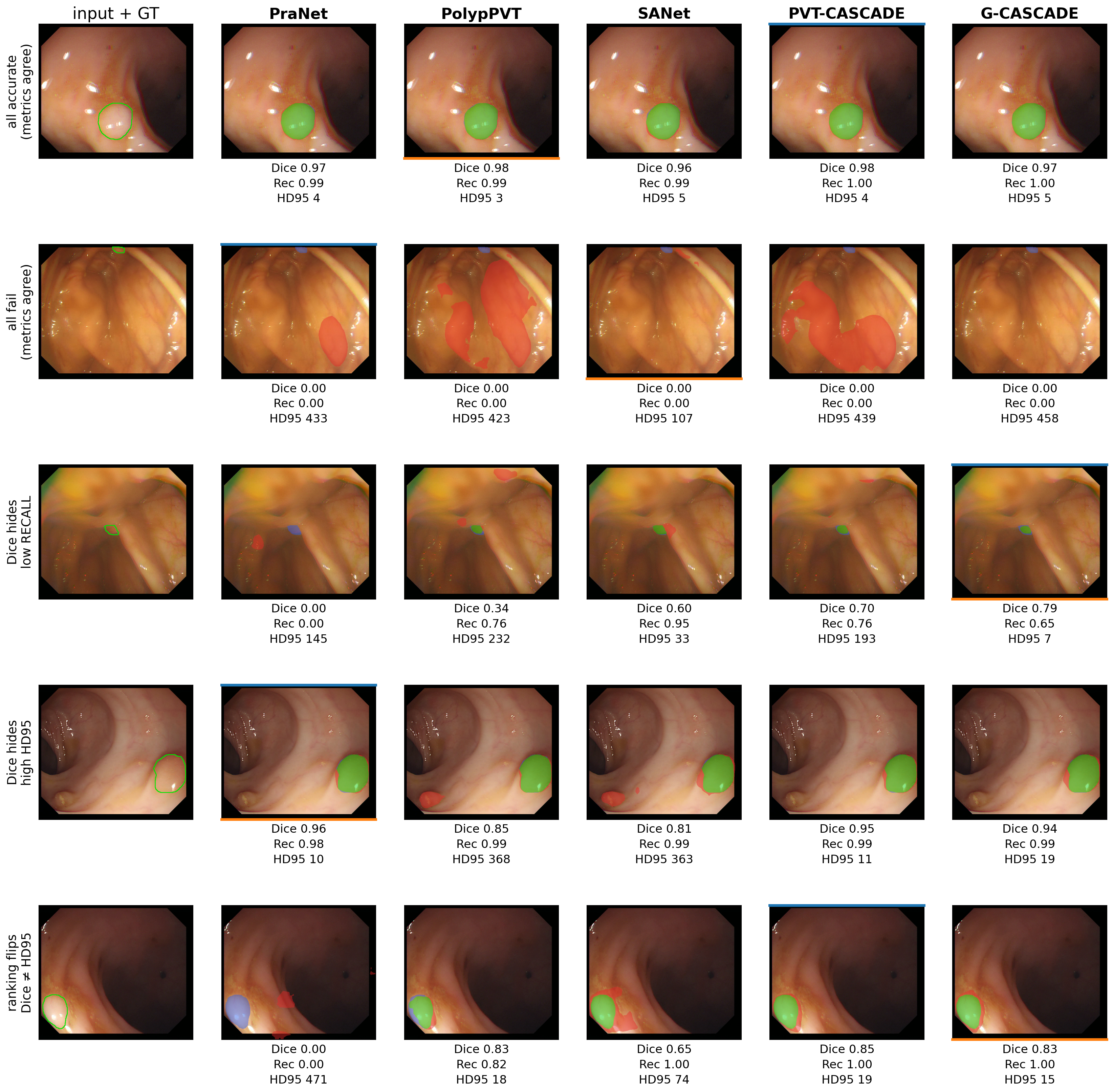}
\vspace{-0.3cm}
\caption{Qualitative case taxonomy on CVC-ColonDB. Columns show the ground truth
and each model name; \textbf{Dice}, recall (\textbf{Rec}) and
\textbf{HD95} are reported beneath every image from columns 1-5. Overlays mark true positives
(green), false positives (red) and false negatives (blue); a blue top border
flags the best-Dice model and an orange bottom border the best-HD95 model. Rows:
(1)~all models accurate -- metrics concur; (2)~all fail; (3)~Dice over-promises
\emph{recall} (missed tissue); (4)~Dice over-promises \emph{HD95} (spurious
far-away FP); (5)~the best-Dice model is not the best-HD95 model.}
\label{fig:taxonomy}
\vspace{-0.4cm}
\end{figure}

\subsection{Results}

\begin{table}[h]
\centering\small
\caption{P1 (fixed split) on the three unseen datasets. Dice is the reported
metric; HD95 (px, $\downarrow$) and Recall ($\uparrow$) are omitted by the
original papers. The model ranking \emph{reshuffles across datasets}, and HD95
exposes failures far larger than the Dice gap implies.}
\vspace{-0.2cm}
\label{tab:p1}
\begin{tabular}{l ccc ccc ccc}
\toprule
& \multicolumn{3}{c}{CVC-300} & \multicolumn{3}{c}{CVC-ColonDB} & \multicolumn{3}{c}{ETIS} \\
\cmidrule(lr){2-4}\cmidrule(lr){5-7}\cmidrule(lr){8-10}
Model & Dice & Rec & HD95 & Dice & Rec & HD95 & Dice & Rec & HD95 \\
\midrule
PraNet      & .877 & .942 & 32.5 & .715 & .730 & 83.2 & .637 & .673 & 230.1 \\
Polyp-PVT   & \textbf{.904} & .947 & \textbf{20.8} & \textbf{.811} & .821 & \textbf{54.8} & \textbf{.790} & .868 & \textbf{126.8} \\
SANet       & .899 & \textbf{.966} & 23.9 & .758 & .798 & 70.4 & .763 & \textbf{.893} & 141.8 \\
PVT-CASCADE & .876 & .955 & 28.4 & .801 & .843 & 59.8 & .771 & .883 & 140.5 \\
G-CASCADE   & .869 & .922 & 23.8 & .703 & .705 & 59.5 & .725 & .765 & 130.1 \\
\bottomrule
\end{tabular}
\vspace{-0.2cm}
\end{table}

\noindent \textbf{P1: region metrics hide boundary and recall failures.}
Three observations follow (Table~\ref{tab:p1}). (i)~\textbf{Boundary failures are
hidden.} On ETIS \cite{silva2014toward}, PraNet's HD95 ($230$\,px) is nearly double Polyp-PVT's
($127$\,px), a gap far larger than the $0.15$ Dice difference suggests. Its recall ($0.67$) reveals that a third of the polyp tissue is missed. (ii)~\textbf{The
ranking is dataset-dependent:} SANet~\cite{sanet2021} ranks 2nd on CVC-300~\cite{vazquez2017benchmark} but 3rd on
CVC-ColonDB~\cite{bernal2015wm}, and ETIS~\cite{silva2014toward}, while PVT-CASCADE~\cite{pvtcascade2023} ranks last on CVC-300~\cite{vazquez2017benchmark} yet 2nd on the harder
sets. (iii)~On the small CVC-300~\cite{vazquez2017benchmark} set ($n{=}60$), pairwise differences between the
top models are \emph{not} statistically significant under FDR-corrected Wilcoxon
tests ($p>0.1$); significance appears only on the larger CVC-ColonDB/ETIS.

\begin{table}[h]
\centering\small
\caption{P2: retraining under three random $80/20$ splits (mean$\pm$std over
seeds). Models lie within $\approx$1\,std of one another and the
PraNet/Polyp-PVT ordering \emph{flips across splits} (seed-1 vs.\ seed-0/2).}
\vspace{-0.2cm}
\label{tab:p2}
\begin{tabular}{lcc}
\toprule
Model & mDice & mHD95 (px) \\
\midrule
HSNet     & $0.930\pm0.005$ & $25.2\pm2.5$ \\
Polyp-PVT & $0.923\pm0.005$ & $27.9\pm1.5$ \\
PraNet    & $0.919\pm0.007$ & $28.9\pm1.9$ \\
\bottomrule
\end{tabular}
\vspace{-0.4cm}
\end{table}

\noindent \textbf{ P2: rankings are not robust to the data split.}
Under random splits (Table~\ref{tab:p2}) the three models are statistically
near-indistinguishable ($\pm0.005$--$0.007$ Dice), and the PraNet vs.\ Polyp-PVT
ranking reverses between seeds. A single fixed-split leaderboard therefore does
not establish a robust ordering -- empirically confirming, on the models
themselves, the split-variance concern raised in our audit.

\begin{table}[h]
\centering\small
\caption{Lesion-level detection (connected-component matching), pooled over the
P1 unseen sets, under three criteria. F1 is in \textbf{bold}. The ranking is
\emph{robust to the criterion} and differs from the pixel-Dice ranking:
G-CASCADE has the best precision (fewest false-alarm lesions), Polyp-PVT the
highest sensitivity, and PVT-CASCADE the best F1; PraNet is consistently worst.}
\vspace{-0.2cm}
\label{tab:lesion}
\begin{tabular}{l ccc ccc ccc}
\toprule
& \multicolumn{3}{c}{any-overlap} & \multicolumn{3}{c}{centroid-in-GT} & \multicolumn{3}{c}{IoU$\geq$0.5} \\
\cmidrule(lr){2-4}\cmidrule(lr){5-7}\cmidrule(lr){8-10}
Model & Sens & Prec & F1 & Sens & Prec & F1 & Sens & Prec & F1 \\
\midrule
PraNet      & .806 & .615 & \textbf{.698} & .757 & .581 & \textbf{.658} & .662 & .501 & \textbf{.571} \\
Polyp-PVT   & \textbf{.898} & .754 & \textbf{.820} & \textbf{.855} & .720 & \textbf{.782} & \textbf{.792} & .640 & \textbf{.708} \\
SANet       & .874 & .802 & \textbf{.836} & .833 & .766 & \textbf{.798} & .739 & .666 & \textbf{.701} \\
PVT-CASCADE & .888 & .849 & \textbf{.868} & .832 & .802 & \textbf{.817} & .782 & \textbf{.748} & \textbf{.765} \\
G-CASCADE   & .819 & \textbf{.891} & \textbf{.854} & .792 & \textbf{.870} & \textbf{.829} & .689 & .741 & \textbf{.714} \\
\bottomrule
\end{tabular}
\vspace{-0.2cm}
\end{table}

\noindent \textbf{Lesion-level detection: a different, clinically-grounded ranking.}
Pixel-Dice rewards overlap but ignores the clinical question -- \emph{did we
detect each lesion without false alarms?} Table~\ref{tab:lesion} reranks the
models: PVT-CASCADE~\cite{pvtcascade2023} attains the best lesion-F1 under every criterion, G-CASCADE~\cite{gcascade2024}
the best precision (fewest false-alarm lesions), and Polyp-PVT~\cite{polypvt2023} the highest
sensitivity. PraNet, competitive on Dice, misses up to a third of lesions and
raises the most false alarms. The conclusion is robust to the matching threshold
(IoU $0.1$--$0.75$).

\begin{table}[h]
\centering\small
\caption{P3 (out-of-distribution): pooled over PolypGen's six unseen centres
($n{=}1537$). All models degrade sharply in absolute terms; the coarse ranking is
largely preserved (Polyp-PVT best, PraNet worst). Best per column in \textbf{bold}.}
\label{tab:p3}
\begin{tabular}{lcccc}
\toprule
Model & Dice & Recall & HD95 & NSD \\
\midrule
Polyp-PVT   & \textbf{0.760} & 0.877 & 146.5 & \textbf{0.339} \\
PVT-CASCADE & 0.758 & \textbf{0.894} & 153.7 & 0.326 \\
SANet       & 0.750 & 0.856 & \textbf{144.9} & 0.260 \\
G-CASCADE   & 0.747 & 0.846 & 152.9 & 0.276 \\
PraNet      & 0.713 & 0.846 & 162.3 & 0.290 \\
\bottomrule
\end{tabular}
\vspace{-0.2cm}
\end{table}
\noindent \textbf{P3: out-of-distribution testing degrades all models.}
On PolypGen's six unseen centres every model drops to Dice $0.71$--$0.76$ with
HD95 $145$--$162$\,px (Table~\ref{tab:p3}), and per-centre Dice ranges from $0.45$
(centre~4) to $0.89$ (centre~3). The coarse ranking is largely preserved, so the
out-of-distribution contribution is not a reshuffle but a uniform, large
\emph{absolute} degradation invisible to in-distribution leaderboards, underscoring why at least one external test set is mandatory.

\noindent \textbf{Qualitative analysis.}
Figure~\ref{fig:taxonomy} shows that the added metrics \emph{agree} with Dice in
concordant cases (rows 1--2; they are not noise) yet \emph{reveal} the failures
Dice hides (rows 3--5). Figure~\ref{fig:explainer} makes the metrics concrete on
a single case.

\section{The Polyp Segmentation Reporting Checklist (PSRC)}
\label{sec:checklist}

Based on the audit findings and the Metrics Reloaded
framework~\cite{metricsreloaded2024}, we propose the following
five-point checklist for future work.
\vspace{-0.3cm}
\begin{enumerate}
  \item[\textbf{C1.}] \textbf{Report HD95 alongside Dice.}
    For any segmentation task where boundary accuracy is clinically
    relevant (lesion sizing, resection margin planning),
    HD95 must be reported.
    Dice alone is insufficient.

  \item[\textbf{C2.}] \textbf{Report Recall / Sensitivity.}
    In screening applications, false negatives carry higher clinical
    cost than false positives.
    Recall must be a primary metric, not an optional supplement.

  \item[\textbf{C3.}] \textbf{Specify split protocol precisely and
    justify comparability.}
    Papers must state whether they use the PraNet fixed split,
    a random split, or a dataset-official split.
    Papers using non-PraNet splits \emph{must not} place their
    numbers in columns that implicitly compare to PraNet-split results.

  \item[\textbf{C4.}] \textbf{Include at least one external or
    OOD test set from an unseen centre.}
    Results on PolypGen, BKAI-IGH, or a private clinical dataset
    are required to support generalisability claims.

  \item[\textbf{C5.}] \textbf{Report statistical significance.}
    Wilcoxon signed-rank or Friedman tests over per-image score
    distributions are sufficient.
    Tabular Dice comparisons with fewer than 2 decimal places
    of separation should not be labelled SOTA without $p < 0.05$.
\end{enumerate}
\section{Discussion and Limitations}
\label{sec:discussion}

\noindent \textbf{Threats to validity.}
Our conclusions rest on several validated assumptions. Self-trained
PVT-CASCADE~\cite{pvtcascade2023} and G-CASCADE~\cite{gcascade2024} reproduce their published in-distribution Dice
within $0.011$ and $0.018$--$0.035$ respectively, confirming faithful training,
so their lower unseen-set scores reflect genuine generalisation rather than
under-training; we nonetheless anchor our metric-disagreement findings on the
released-weight models wherever possible. P1 self-trained results are
single-seed, but our own P2 experiment directly quantifies training-seed
variance ($\pm0.005$--$0.007$ Dice), which we fold into our interpretation of
near-tied rankings. 
HD95 is reported under the \texttt{medpy} convention and
differs from the Metrics-Reloaded max-of-directed convention by up to
$\sim16$\, px on identical masks, which itself is a reproducibility finding. Therefore, absolute HD95 values should be compared only within a fixed implementation.
Finally, a perceptual-hash$\rightarrow$SSIM audit confirms the P1 train/test split is disjoint (max cross-set SSIM $0.66$); we did not additionally SSIM-audit the independent-institution PolypGen OOD centres against the training pool, which we note as a residual low risk. Extended reproduction numbers, split-dependence clarifications, and lesion-matching robustness checks are provided in the supplementary material.

\noindent \textbf{Discussion.}
The field's failure to adopt surface-distance metrics like HD95 despite an
architectural focus on edge refinement stems from benchmarking path
dependency. The PraNet template, established in 2020 for rapid baseline
comparison, predated the 2021--2022 wave of boundary-focused architectures.
Because subsequent papers copied these exact tabular layouts to preserve
direct leaderboard comparability, the evaluation criteria remained
artificially frozen while the underlying scientific questions evolved.

Our audit findings represent a conservative lower bound on these reporting
vulnerabilities, as we relied strictly on the main experimental sections of
published text rather than hidden appendices or external repositories.
Furthermore, this audit is limited to still-image colonoscopic polyp
segmentation; extending this meta-evaluation framework to parallel fields
like skin lesion segmentation remains a direction for future work.
\section{Conclusion}
\label{sec:conclusion}

We systematically audited and empirically stress-tested a decade of evaluation practices in colonoscopy polyp segmentation. Our meta-evaluation of 27 papers reveals that the sub-field relies on fragile benchmarking conventions characterized by widespread metric omissions, incompatible split mixing, a lack of statistical rigor, and pervasive transcription errors. By uniformly re-evaluating five representative models across three controlled protocols, we demonstrated that standard region-overlap metrics silently mask severe boundary and false-negative failures, and that reported leaderboard rankings collapse under stochastically altered training splits. To correct this path dependency, we introduce the Polyp Segmentation Reporting Checklist (PSRC) alongside an open-source evaluation toolkit. Ultimately, our findings serve as a broader warning to the medical imaging community that benchmarking protocols, rather than architectural innovation alone, dictate reported progress; the released toolkit provides an immediate path toward standardized, robust, and clinically accountable evaluation.

\bibliographystyle{splncs04}   
\bibliography{references}

\newpage

\section{Additional Results}

\begin{figure}[h]
\centering
\includegraphics[width=0.89\linewidth]{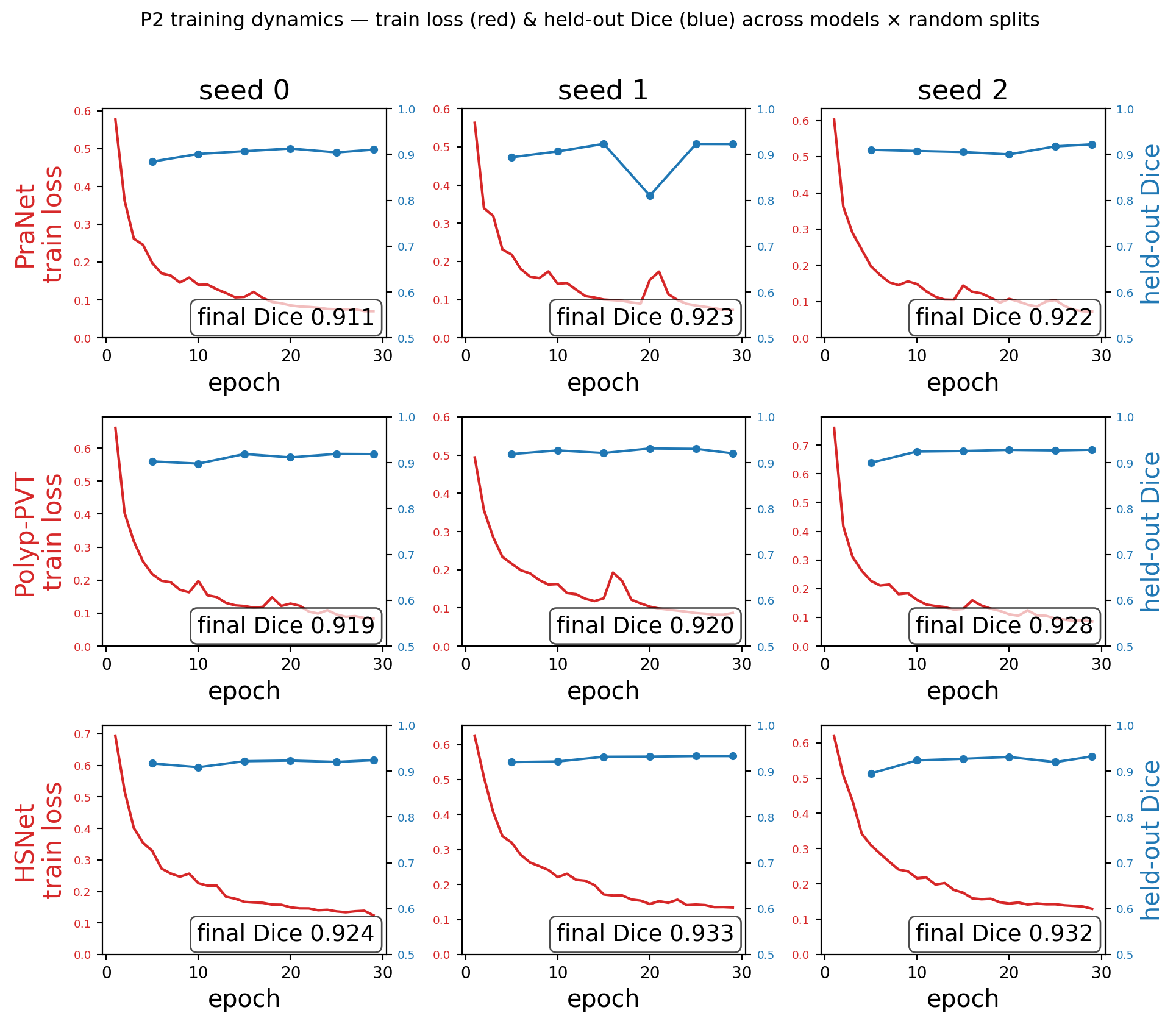}
\caption{P2 training curves (rows: models; columns: random splits): train loss
(red) and held-out Dice (blue). Held-out Dice plateaus by $\sim$5 epochs while
train loss keeps falling (mild memorisation, no degradation); curves are nearly
identical across seeds.}
\label{fig:loss}
\vspace{-0.4cm}
\end{figure}

\noindent \textbf{Training dynamics (P2).}
Held-out Dice converges within $\sim$5 epochs and plateaus while train loss
continues to fall, indicating diminishing returns rather than harmful
over-fitting (Fig.~\ref{fig:loss}). Notably, the transformer HSNet retains a
higher train loss ($0.13$) yet the best held-out Dice ($0.93$), whereas the CNN
PraNet fits the training data hardest ($0.07$) but generalises slightly worse, and the dynamics are stable across random splits.

\section{Threats to Validity}
\label{sec:threats}

\noindent \textbf{Released vs.\ self-trained weights.}
PraNet~\cite{pranet2020}, Polyp-PVT~\cite{polypvt2023}, and SANet~\cite{sanet2021} are evaluated from the authors' \emph{released}
checkpoints; PVT-CASCADE~\cite{pvtcascade2023} and G-CASCADE~\cite{gcascade2024} have no released polyp weights and are
\emph{self-trained} on the standard 1450-image split. To show this does not
confound our conclusions, we verify in-distribution reproduction on the held-out
Kvasir-SEG~\cite{jha2019kvasir} and CVC-ClinicDB~\cite{bernal2015wm} test sets: PVT-CASCADE~\cite{pvtcascade2023} matches its published Dice
almost exactly (Kvasir $0.917$ vs.\ $0.917$; ClinicDB $0.932$ vs.\ $0.943$), and
G-CASCADE reproduces within $0.02$--$0.035$ (Kvasir $0.893$ vs.\ $0.927$; ClinicDB
$0.929$ vs.\ $0.947$). The self-trained models are therefore faithful, and their
lower scores on the \emph{unseen} sets reflect genuine generalisation behaviour,
not under-training. Nonetheless, we anchor the metric-disagreement findings on
the released-weight models wherever possible, and treat self-trained absolute
scores as secondary.

\noindent \textbf{What is (and is not) protocol-dependent.}
We are deliberately precise: the fixed-split (P1) and out-of-distribution (P3)
rankings are in fact largely \emph{consistent} (Polyp-PVT best, PraNet worst on
both), and P3's contribution is to show uniform \emph{absolute} degradation on
unseen centres (Dice $0.71$--$0.76$, HD95 $145$--$162$\,px) rather than a
reshuffle. The instability we document is narrower and better supported: (i)
under three random $80/20$ splits (P2) adjacent, near-tied models reorder
(PraNet$\leftrightarrow$Polyp-PVT), and (ii) \emph{metric choice} reorders the
podium (G-CASCADE ranks 4th on Dice but 1st on lesion precision). We therefore
claim that near-tied, single-metric, single-split leaderboard gaps are not
robust -- not that architecture rankings are arbitrary.

\noindent \textbf{Single-seed training and statistical power.}
Self-trained P1 models are single runs; our own P2 quantifies the training-seed
variance ($\pm0.005$--$0.007$ Dice), which we fold into the interpretation.
Significance tests on the small benchmarks are correspondingly low-powered
(CVC-300, $n{=}60$); we therefore report bootstrap confidence intervals and
rank-biserial effect sizes alongside FDR-corrected $p$-values, and frame results
as ``gaps within noise'' rather than proven equivalence.

\noindent \textbf{Metric-implementation scope.}
Our Dice, IoU, recall, and NSD match the official \texttt{MetricsReloaded}
reference to $\leq10^{-11}$ (NSD exactly). HD95 is reported under the medpy
convention and differs from the Metrics-Reloaded max-of-directed convention by up
to $\sim\!16$\,px on the same masks -- itself a reproducibility finding, but it
means absolute HD95 values should be compared only within a fixed implementation.
The Metrics-Reloaded \emph{recommendation} we adopt follows the published problem
fingerprint; confirming it via the interactive toolkit is left as a reporting
step.

\noindent \textbf{Lesion-detection robustness.}
Connected-component matching currently applies no minimum-size filter, so a
single-pixel false-positive speck can count as a false-alarm lesion. We report
the metric under three matching criteria (any-overlap, centroid-in-GT,
IoU$\geq$0.5) and across IoU thresholds $0.1$--$0.75$; the model ordering is
stable across all of them. Adding a small (e.g.\ $\geq$0.1\% area) component
filter is recommended for deployment and does not change the ranking.

\noindent \textbf{Leakage.}
A perceptual-hash$\rightarrow$SSIM audit confirms the P1 train/test split is
disjoint (max cross-set SSIM $0.66$). The PolypGen OOD centres are from
independent institutions; we did not additionally SSIM-audit PolypGen against the
training pool, which we note as a residual (low) risk.

\end{document}